\documentclass[preprint,12pt]{elsarticle}
\usepackage{lineno,hyperref}
\usepackage{colortbl}
\usepackage{graphicx}
\usepackage{float}
\usepackage{amsmath}
\usepackage{booktabs}
\usepackage{amsfonts}
\usepackage{algorithm}
\usepackage{algorithmic}
\usepackage{amssymb}
\usepackage{multirow}
\usepackage{bm}

\journal{Information Fusion}

\begin{document}

\begin{frontmatter}

\title{Learnable Graph Convolutional Network and Feature Fusion for Multi-view Learning}

\author[label1,label2]{Zhaoliang~Chen}
\author[label1,label2]{Lele~Fu}
\author[label1,label2]{Jie~Yao}
\author[label1,label2]{Wenzhong~Guo}
\author[label3,label4]{Claudia~Plant}
\author[label1,label2]{Shiping~Wang\corref{cor1}}
\ead{shipingwangphd@163.com}
\cortext[cor1]{Corresponding author.}

\address[label1]{College of Computer and Data Science, Fuzhou University, Fuzhou 350116, China}
\address[label2]{Fujian Provincial Key Laboratory of Network Computing and Intelligent Information Processing, Fuzhou University, Fuzhou 350116, China}
\address[label3]{Faculty of Computer Science, University of Vienna, Vienna 1090, Austria}
\address[label4]{Research Network Data Science @ Uni Vienna, University of Vienna, Vienna 1090, Austria}





\begin{abstract}
    In practical applications, multi-view data depicting objectives from assorted perspectives can facilitate the accuracy increase of learning algorithms.
    However, given multi-view data, there is limited work for learning discriminative node relationships and graph information simultaneously via graph convolutional network that has drawn the attention from considerable researchers in recent years.
    Most of existing methods only consider the weighted sum of adjacency matrices, yet a joint neural network of both feature and graph fusion is still under-explored.
    To cope with these issues, this paper proposes a joint deep learning framework called Learnable Graph Convolutional Network and Feature Fusion (LGCN-FF), consisting of two stages:
    feature fusion network and learnable graph convolutional network.
    The former aims to learn an underlying feature representation from heterogeneous views,
    while the latter explores a more discriminative graph fusion via learnable weights and a parametric activation function dubbed Differentiable Shrinkage Activation (DSA) function.
    The proposed LGCN-FF is validated to be superior to various state-of-the-art methods in multi-view semi-supervised classification.

\end{abstract}

\begin{keyword}
    Information fusion with deep learning, multi-view learning, graph convolutional network, semi-supervised classification.
\end{keyword}

\end{frontmatter}

\section{Introduction}
\begin{figure}[!htbp]
  \centering
  \includegraphics[width=0.75\textwidth]{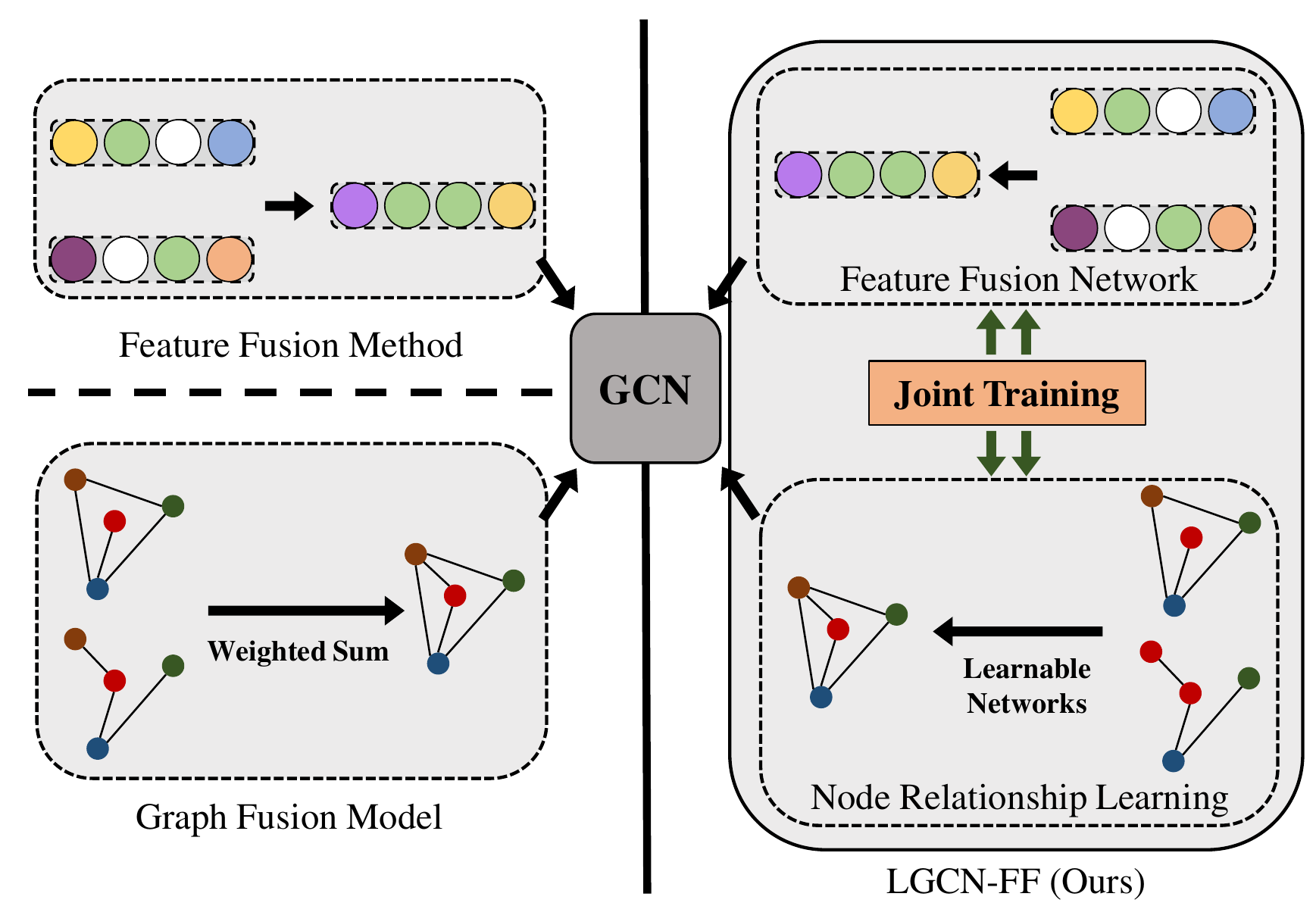}
  \caption{A brief comparison of the proposed LGCN-FF and other methods.
  Compared with most multi-view learning models (left) that only concentrated on either feature fusion approaches or graph fusion approaches, LGCN-FF (right) is a joint framework that simultaneously trains feature fusion networks and learnable graph fusion networks with a multi-stage training strategy.
  }
  \label{MethodCompare}
\end{figure}

In real-world applications, a large amount of information exists in varied forms, because an object can be described from heterogeneous data sources.
For example, streaming media can be illustrated by features of frames, audio, and textual descriptions, which come into being multi-view data.
This motivates researchers to discover the latent consistent information across diverse views \cite{chen2021relaxed,wang2022hyperspectral,ChenWPHZ21}.
Instead of directly exploiting features from heterogeneous sources, it should be helpful to extract node relationships among samples and propagate supervision signals across nodes,
which motivates us to conduct graph learning on multi-view data.
Graph learning is a crucial field of machine learning in decades and has been extensively applied to a multitude of practical applications,
such as node classification \cite{TangZLLWZW19,KipfW17,yao2019graph}, social network analysis \cite{YanhuiHybrid2022,BaiCJRH22, RawatSK18} and computer vision \cite{ChenXPLZ22,wang2021covid,mosella20212d}.
In recent years, Graph Convolutional Network (GCN) has been widely explored for its powerful ability to integrate the connectivity patterns and feature attributes with given graph-structural data \cite{KipfW17}.
A large number of studies have revealed the remarkable performance boosting of GCN in various learning tasks \cite{zhang2021shne,ZhouSLLNYS20,zhu2022interpretable}.
Although most of multi-view datasets do not naturally contain topological structure like traditional citation and link prediction datasets, samples in real-world applications often have implicit connections that can be extracted.
In light of this, we can mine the hidden relationships among samples via existing features and generate multifarious graphs.
These estimated graphs generally describe node relationships from various perspectives with their complementarity.
As is shown on the left side of Figure \ref{MethodCompare}, most existing methods utilize feature fusion methods or graph fusion models before applying GCN, both of which are critical for the performance of downstream multi-view learning tasks.

However, limited work has concentrated on co-optimization of feature and graph fusion learning via GCN.
As we have discussed, multi-view data in the real-world generally do not exist as the network topology, attributed to which most algorithms preprocess the original data and convert them to graph-structural data.
Because $k$-nearest neighbor (KNN) can estimate edges among nodes via calculating similarities of samples and exploring nearest neighbors,
most of the existing methods generated graphs via KNN.
However, some studies have pointed out that this is often inaccurate and may yield undesired links between samples \cite{li2021consensus}.
Previous works generally applied these adjacency matrices without training or refining on neighborhood relationships \cite{LiLW20a,0017ZB0SP20,HuangZZZLP21}, which possibly resulted in performance decline of multi-view learning.
Although some researchers have successfully leveraged GCN to deal with multi-view data \cite{LiLW20a,ChengWTXG20}, they only consider a weighted combination of different adjacency matrices.
This is problematic because a linear weighted sum of adjacency matrices may amplify and aggregate incongruous noises from distinct graphs estimated by KNN.
Therefore, a well-established graph refining procedure should be conducted after numerous graphs are integrated into a unique one, so that the negative impact of undesired connections is mitigated.

To our knowledge, there is very limited work on the joint training of feature and graph fusion learning, which is beneficial to exploring co-optimal solutions to both two problems.
Consequently, we propose an end-to-end framework dubbed Learnable Graph Convolutional Network and Feature Fusion (LGCN-FF).
A brief description is shown on the right side of Figure \ref{MethodCompare}.
LGCN-FF is comprised of two fundamental components: a feature fusion network and a learnable GCN network.
The former aims to resolve the feature fusion problem with given multi-view data, and the latter is to learn adjacency matrix fusion with multiple graphs generated from distinct features.
Feature fusion is realized by multiple sparse autoencoders and a fully-connected network that is responsible for incorporating all features.
Adjacency matrix fusion is first conducted by a weighted sum of all graphs, where all weights are learned automatically.
For the purpose of learning a more discriminative graph representation, we present a learnable function termed as Differentiable Shrinkage Activation (DSA) to further explore adjacency matrix fusion,
which adaptively refines feasible and robust node relationships during training.
It can be regarded as an analogous pattern to soft thresholding operator in Iterative Shrinkage Thresholding Algorithm (ISTA) \cite{DBLP:conf/icml/GregorL10} and Singular Value Thresholding (SVT) \cite{CaiCandes10Singular}, which are applied to address sparse coding and low-rank approximation problems, respectively.
Each iteration of the proposed LGCN-FF contains four optimization steps in light of their own loss functions.
Therefore, LGCN-FF is a joint framework that learns features and node relationships simultaneously.
The main contributions of this paper are as follows:

(1) Propose an end-to-end neural network framework for multi-view semi-supervised classification, which integrates sparse autoencoders and a learnable GCN to jointly learn intact representations of multiple features and graphs.

(2) Construct a learnable GCN framework with adaptive weights and a parametric DSA function, both of which mine more discriminative and robust representations of graphs from heterogeneous views automatically.

(3) Develop a multi-step optimization strategy for LGCN-FF via back propagation, each of which updates corresponding parameters while fixing other learnable parameters.

(4) The proposed framework is leveraged to conduct multi-view semi-supervised classification tasks, and achieves superior performance compared with other state-of-the-art graph-based algorithms.

The rest of this paper is organized as follows.
Related works on GCN, multi-view learning, feature and graph fusion are reviewed in Section \ref{RelatedWork}.
We elaborate the proposed LGCN-FF in Section \ref{ProposedMethod}, including the detailed introduction of each component and algorithm analyses.
Finally, the effectiveness of the proposed framework is verified via substantial experiments in Section \ref{Experiment}, and our work is concluded in Section \ref{Conclusion}.

\section{Related Work}\label{RelatedWork}
\subsection{Graph Convolutional Network}
In this subsection, we first review recent works on GCN.
A spectral graph convolution operation is conducted by a signal $x \in \mathbb{R}^{m}$ and a filter $g_{\theta} = diag(\theta)$, formulated as
\begin{gather}\label{OriginalGCN}
g_{\theta} \star x=\mathbf{U} g_{\theta} \mathbf{U}^{\top} x,
\end{gather}
where $\mathbf{U}$ denotes the matrix of eigenvalues of the normalized graph Laplacian matrix.
For the purpose of saving computational resources, Kipf et al. \cite{KipfW17} performed the first-order approximation of truncated Chebyshev polynomial and imposed it on the node classification tasks with network topology.
Specifically, the $l$-th layer of a spectral GCN is formally defined as
\begin{gather}\label{GCN}
\begin{split}
\mathbf{H}^{(l)} = \sigma \left( \tilde{\mathbf{D}}^{-\frac{1}{2}} \tilde{\mathbf{A}} \tilde{\mathbf{D}}^{-\frac{1}{2}} \mathbf{H}^{(l-1)} \mathbf{W}^{(l)} \right),
\end{split}
\end{gather}
where $\tilde{\mathbf{A}} = \mathbf{A} + \mathbf{I}$ denotes the adjacency matrix considering the self-connections, and $[\tilde{\mathbf{D}}]_{ii} = \sum_{j} [\tilde{\mathbf{A}}]_{ij}$.
Layer-specific weight matrix is denoted by $\mathbf{W}^{(l)}$.
The graph convolution operation can be regarded as a special form of Laplacian smoothing \cite{LiHW18}, which propagates the neighborhood features across the whole network topology.
Due to the encouraging performance of GCN, many variant algorithms have been explored.
For example,
Xu et al. put forward an innovative answer-centric approach dubbed radial graph convolutional networks to cope with the visual question generation tasks \cite{XuWYHS21}.
Liu et al. integrated GCN with hidden conditional random field to reserve the skeleton structure information during the classification stage \cite{LiuGKQG21}.
Bo et al. investigated the low-frequency and high-frequency signals in a graph, and proposed a model that adaptively integrated different signals during message passing \cite{BoWSS21}.
A variant of GCN was derived through a modified Markov diffusion kernel, which explored the global and local contexts of nodes \cite{ZhuK21}.
Guo et al. exploited GCN to propagate features over the relationship affinity matrix, generating relationship-regularized representations of objectives to produce the scene graph \cite{guo2021relation}.
A convolution operator on the multi-relational graph was developed, based on which the proposed multi-dimensional convolution operator achieved the eigenvalue decomposition of a Laplacian tensor \cite{MRGCN2020HUANG}.
Lei et al. established the graph receptive fields according to diffusion paths and applied them to build a compact graph convolutional network \cite{lei2020diggcn}.
A multi-stage GCN-based framework was presented with the self-supervised learning to improve the generalization performance on the graph with limited supervised information \cite{SunLZ20}.
These GCN-based works have significantly promoted the performance of various learning tasks in both Euclidean and non-Euclidean domains.

\subsection{Multi-view Learning}
Multi-view learning that leverages assorted types of features from heterogeneous views has promoted the performance of various machine learning tasks \cite{LiuMulti20}.
A multi-view and multi-feature learning framework was constructed to simultaneously consider the fusion of features and views, which refined a discriminant representation from distinctive classes \cite{LI2019Generative}.
Chen et al. proposed a joint framework for multi-view spectral clustering by learning an adaptive transition probability matrix \cite{ChenAda21}.
The nuclear norm-based optimization method was proposed to conduct multi-view image data fusion via a joint learning framework \cite{huang2020oriented}.
Late fusing incomplete multi-view clustering was proposed to learn a cluster assignment from distinct views to exploit a consensus clustering matrix \cite{liu2019late}.
A$\rm{E}^2$-Nets utilized inner autoencoders to perform view-specific representation learning, and adopted the outer autoencoders to implement multi-view information encoding \cite{Zhang2019AE}.
Wen et al. presented an effective incomplete multi-view clustering framework to make full use of the local geometric information and the unbalanced discriminating powers of incomplete multi-view observations \cite{WenZZFW21}.
All of these works have validated the encouraging performance of multi-view learning compared with single-view learning.

\subsection{Feature or Graph Fusion}
Effective feature or graph fusion is universally applied to the multi-view data processing to achieve desired learning performance, which takes advantage of full observations from multi-view representations.
Zhou et al. fused information from multiple kernels to improve the performance of multi-kernel clustering \cite{zhou2020subspace}.
Tang et al. proposed a deep neural network that recurrently fused and refined multi-scale deep features \cite{TangLZLXWZL22}.
Graphs of multiple views can be integrated into a consistent global graph, whose Laplacian matrix is constrained with multiple strongly connected components \cite{NieLiLi17SelfWeighted}.
Huang et al. put forward a unified multi-view image data fusion model on the basis of nuclear norm optimization \cite{HuangZL20}.
\cite{NieCaiLi17Multiview} paid attention to preserving the local structure of data while conducting graph fusion.
A graph neural network-based fusion mechanism was designed to extract complementary information across views \cite{He2020MVGNN}.
\cite{Wang2020GMC} learned graph matrices of heterogeneous views, and a unified graph matrix is recovered via a mutual reinforcement manner.
A unified framework was proposed by introducing a co-training strategy into the GCN framework, where the graph information embedded in multiple views is explored adaptively \cite{LiLW20a}.
Nonetheless, most of them only concentrate on either graph fusion or feature fusion, both of which influence the performance of GCN considerably.
It is pivotal to develop a framework with a co-training pattern that can conduct feature fusion and graph fusion simultaneously.
In the following section, we will elaborate the proposed method to solve this issue, being the main contribution of this paper.

\section{The Proposed Method}\label{ProposedMethod}

\begin{figure}[!htbp]
    \centering
    \includegraphics[width=0.9\textwidth]{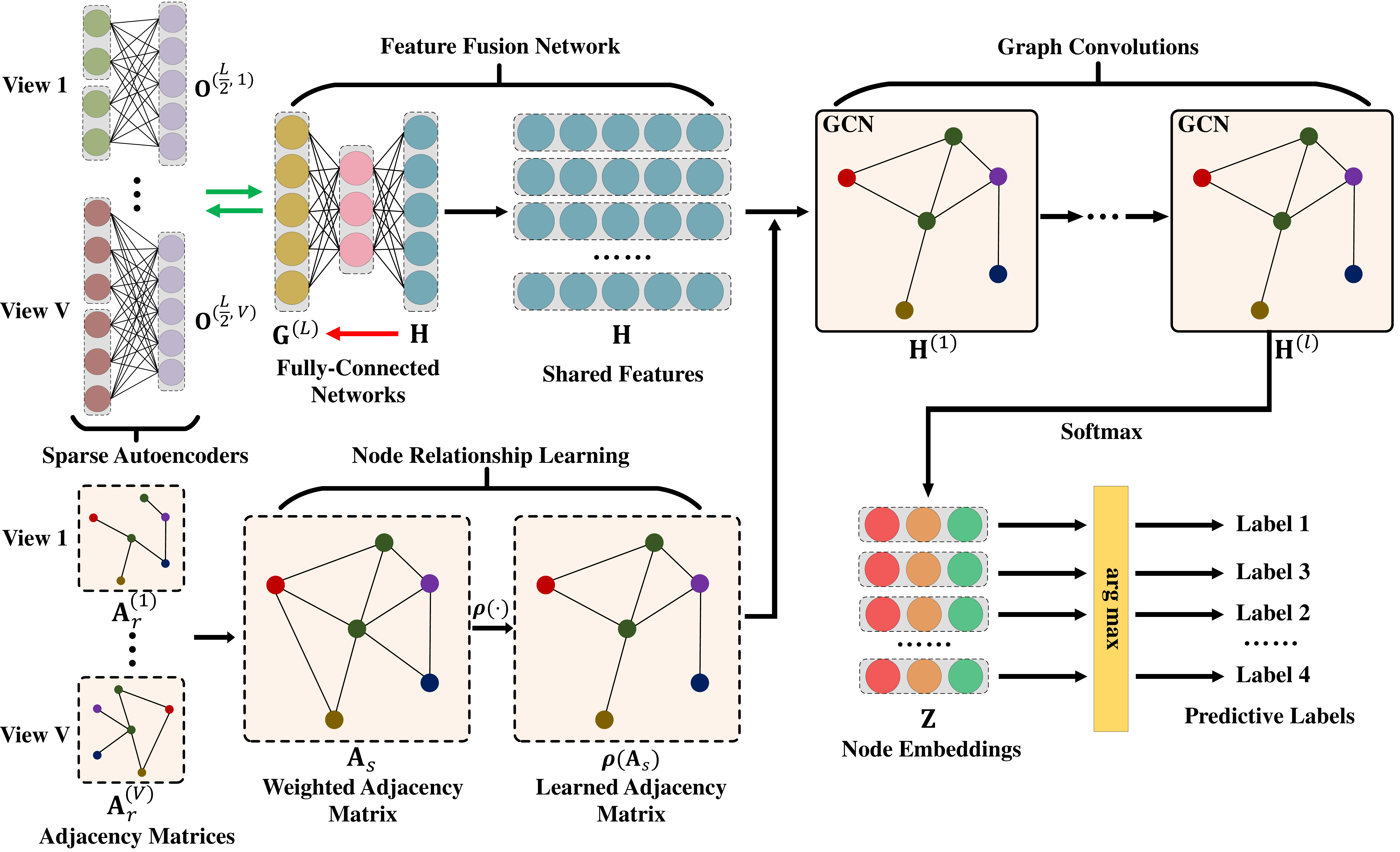}
    \caption{Structure of the proposed LGCN-FF, which consists of a feature fusion network and a learnable GCN.
    LGCN-FF is an end-to-end framework whose learnable parameters are updated by a multi-step optimization strategy.
    Intact representations of both multi-view features and graphs are learned simultaneously to promote the performance of GCN.
    }
    \label{Framework}
  \end{figure}

For the purpose of jointly learning feature fusion and graph fusion, we develop an end-to-end unified neural network framework consisting of two primary components:
feature fusion network and learnable GCN.
The optimization procedure of one iteration in this framework is divided into multiple steps, inspired by the Alternating Direction Minimization (ADM) \cite{LinLS11} strategy.
In particular, each independent optimization step has its own loss function, all of which make up a complete training iteration of the whole network.
Given $m$ samples with $n$ features, the proposed LGCN-FF aims to solve semi-supervised classification problems with given multi-view data $\mathcal{X} = \{\mathbf{X}^{(1)}, \cdots, \mathbf{X}^{(V)} \}$,
where $\mathbf{X}^{(v)} \in \mathbb{R}^{m \times n_{v}}$ denotes the features of the $v$-th view with totally $V$ views.
Figure \ref{Framework} provides a detailed illustration of the proposed LGCN-FF.
Feature fusion aims to integrate multi-view features with varying dimensions into an intact representation with the same dimension, exploring the underlying features.
Learnable GCN is supposed to merge multiple adjacency matrices, generating a unique graph with better robustness and generalization.

\subsection{Feature Fusion Network}
In order to tackle multi-view data consisting of various features with varied dimensions and explore the underlying information across multiple views,
the proposed LGCN-FF firstly projects original multi-view representations onto a shared latent space.
Considering that feature dimensions of multiple perspectives may extremely differ,
it is not applicable to mapping these features onto the same compressed latent space.
Therefore, we employ sparse autoencoders to explore overcomplete underlying representations for all views.
Each view corresponds to a view-specific sparse autoencoder, and these sparse autoencoders map the original features onto the sparse space with the same dimension.
Formally, the output $\mathbf{O}^{(l, v)} \in \mathbb{R}^{m \times d_{l}}$ of the $l$-th layer for the $v$-th sparse autoencoder is defined as 
\begin{gather}\label{SAE}
\begin{split}
\mathbf{O}^{(l, v)} = \sigma \left(  \mathbf{O}^{(l-1, v)} \mathbf{W}_{sa}^{(l, v)} + \mathbf{b}_{sa}^{(l, v)} \right),
\end{split}
\end{gather}
where $\mathbf{W}_{sa}^{(l, v)} \in \mathbb{R}^{d_{l-1} \times d_{l}}$ and $\mathbf{b}_{sa}^{(l, v)} \in \mathbb{R}^{d_{l}}$ are layer-specific weight and bias, respectively.
The input of the $v$-th sparse autoencoder is $\mathbf{X}^{(v)}$, i.e., $ \mathbf{O}^{(0, v)} = \mathbf{X}^{(v)}$.
Here, $\sigma(\cdot)$ is the layer-specific activation function.
The loss function of the sparse autoencoder in the $v$-th view is written as
\begin{gather}\label{SAELoss}
\begin{split}
\mathcal{L}_{sa}^{(v)} = \frac{1}{2} \left\| \mathbf{O}^{(L, v)} - \mathbf{X}^{(v)} \right\|_{2}^{2} + \beta \mathcal{D}_{KL}(\rho \| \hat{\rho}),
\end{split}
\end{gather}
where $\mathcal{D}_{KL}$ is the Kullback-Leibler divergence and $\beta$ controls the sparsity penalty degree.
Hyperparameter $\rho$ is the value maintaining the sparsity, and $\hat{\rho}$ is the average of the distribution of latent neuron activations.
Kullback-Leibler divergence is calculated by
\begin{gather}\label{KL}
\begin{split}
\mathcal{D}_{KL}(\rho \| \hat{\rho}) = \rho \mathrm{log} \frac{\rho}{\hat{\rho}} + (1 - \rho) \mathrm{log} \frac{(1 - \rho)}{(1 - \hat{\rho})}.
\end{split}
\end{gather}

In order to fuse the hidden features from diverse perspectives into a shared feature vector, we further utilize a fully-connected neural network to carry out the feature fusion task.
Assuming that there are a total of $L$ layers in the fully-connected neural network, the forward propagation in the $l$-th layer is computed by
\begin{gather}\label{FCNet}
\begin{split}
\mathbf{G}^{(l)} = \sigma \left(  \mathbf{G}^{(l-1)} \mathbf{W}_{fc}^{(l)} + \mathbf{b}_{fc}^{(l)} \right),
\end{split}
\end{gather}
where $\mathbf{G}^{(0)} = \mathbf{H}$.
The matrix $\mathbf{H}$ is also a learnable input updated by gradient descent and back propagation techniques.
We project the learned $\mathbf{H}$ onto various view-specific latent features $\{ \mathbf{O}^{(\frac{L}{2}, v)} \}_{v=1}^{V}$ via a trainable fully-connected network.
In fact, the trainable $\mathbf{H}$ also serves as the shared node representation in the learnable GCN.
A two-step optimization strategy is employed for updating $\{ \mathbf{W}_{fc}^{(l)},  \mathbf{b}_{fc}^{(l)}\}_{l = 1}^{L}$ and $\mathbf{H}$.
Both two steps share the same reconstruction loss function as defined below:
\begin{gather}\label{FCLoss}
\begin{split}
\mathcal{L}_{fc} = \frac{1}{2}  \sum_{v=1}^{V} \left\| \mathbf{G}^{(L)} - \mathbf{O}^{(\frac{L}{2}, v)} \right\|_{2}^{2}.
\end{split}
\end{gather}
With the assumption that features of each single view can be rebuilt from the intact common representation $\mathbf{H}$ by trainable weights and biases in the fully-connected network,
Equation \eqref{FCLoss} is regarded as the trade-off of reconstruction errors among heterogeneous views and explores the shared underlying features.
We present the details of optimization steps in Section \ref{TrainingStrategy}.

\subsection{Learnable Graph Convolutional Network}
In this subsection, we present a learnable GCN which automatically integrates the adjacency matrices generated by multiple views and learns a graph containing more discriminative node relationships.
Firstly, the adaptive weighted sum of adjacency matrices is obtained by
\begin{gather}\label{WeightedA}
\begin{split}
\mathbf{A}_{s} = \sum_{v=1}^{V} \pi^{(v)} \mathbf{\mathbf{A}}^{(v)}_{r},
\end{split}
\end{gather}
where $\mathbf{\mathbf{A}}^{(v)}_{r}  = (\tilde{\mathbf{D}}^{(v)})^{-\frac{1}{2}} \tilde{\mathbf{A}}^{(v)} (\tilde{\mathbf{D}}^{(v)})^{-\frac{1}{2}}$ is the initial renormalization adjacency matrix of the $v$-th view,
and $\pi^{(v)}$ is the automatically learned view-specific weight coefficient.
The initialization of adjacency matrices can be conducted via the KNN method.
Because $\{ \pi^{(v)} \}_{v=1}^{V}$ is constrained with $\sum_{v=1}^{V} \pi^{(v)} = 1$, we employ the softmax renormalization at each epoch as
\begin{gather}\label{NormalizationPai}
\begin{split}
\pi^{(v)} \leftarrow \frac{\mathrm{exp} \left( \pi^{(v)} \right)}{ \sum_{v=1}^{V} \mathrm{exp} \left( \pi^{(v)} \right) }
\end{split}
\end{gather}
for $v = 1, \cdots, V$.

Nevertheless, a straightforward weighted sum of adjacency matrices may not be sufficiently feasible for multi-view graph learning,
because a linear weighted sum of all graphs may yield undesired connections between nodes in the fused graph.
Besides, attributed to the fact that the neighborhood relationships are estimated by KNN, which may be not accurate enough, a data-driven refining process should be taken to explore a more comprehensive graph fusion without corrupting the structure and characteristic information of the original graphs.
In order to achieve an optimal adjacency matrix fusion for the given task, we propose the Differentiable Shrinkage Activation (DSA) function denoted by $\rho (\cdot)$ to refine the weighted adjacency matrix.
Because GCN is developed with the precondition that the graph should be undirected,
we require that the output of $\rho (\mathbf{A}_{s})$ should also be symmetrical.
To this end, the learnable DSA function $\rho (\cdot)$ is defined as
\begin{gather}\label{ActivationS}
\begin{split}
\rho (\mathbf{A}_{s}) = \mathbf{A}_{s} \odot  \mathrm{ReLU} \left( \mathbf{S} - \mathbf{\Theta}  \right),
\end{split}
\end{gather}
where $\odot$ is the Hadamard product (entry-wise product),
$\mathbf{S} \in \mathbb{R}^{m \times m}$ denotes the learnable coefficient matrix and $\mathbf{\Theta} \in \mathbb{R}^{m \times m}$ controls the thresholds of node relationship activations.
For the sake of theoretic strictness and better interpretation, it is required that $\mathbf{S}$ and $\mathbf{\Theta}$ should be symmetrical.
Therefore, we define the coefficient matrix $\mathbf{S}$ as
\begin{equation}\label{LearnableS}
\mathbf{S} = \mathrm{Sigmoid} \left( \frac{1}{2} \left(\mathbf{\bar{S}} + \mathbf{\bar{S}}^{\mathrm{T}}\right) \right),
\end{equation}
which is parameterized by a learnable matrix $\mathbf{\bar{S}} \in \mathbb{R}^{m \times m}$.
On the basis of Equation \eqref{LearnableS}, the proposed method can learn an edge-specific coefficient for each edge of the undirected graph,
which automatically shrinks node relationships with coefficient values ranging in $[0, 1]$.

In order to reduce local data noises and construct a sparser graph, the learnable matrix $\mathbf{\Theta}$ in Equation \eqref{ActivationS} is considered as a thresholding matrix controlling the edge activation.
For simplicity and theoretical rigor, we define the entry of the thresholding matrix as
\begin{equation}
[ \mathbf{\Theta} ]_{ij} = [ \mathbf{\Theta} ]_{ji} = \mathrm{Sigmoid} (\theta_{i}), \; \forall i \leqslant j \leqslant m
\end{equation}
with $\bm{\theta} = [\theta_{1}, \cdots, \theta_{m}]$, where $\bm{\theta}$ is a learnable vector and $\mathrm{Sigmoid}(\cdot)$ admits the non-negativity of thresholders.
Consequently, $\mathbf{\Theta}$ is symmetrical and promotes the sparseness of outputs calculated by Equation \eqref{ActivationS}, which can also be regarded as trainable biases of the coefficient matrix.
It is noted that only the node relationship information whose coefficient is greater than its corresponding thresholding value can be activated.
DSA function is beneficial for improving the performance of GCN, due to the ability of automatical feature learning via coefficient matrix and thresholding values.
Actually, it is an analogous pattern as shrinkage function widely employed in proximal optimization which promotes the sparse or low-rank property, e.g. the soft thresholding operator in ISTA \cite{DBLP:conf/icml/GregorL10} or SVT \cite{CaiCandes10Singular} algorithms.
However, the classical thresholders are generally hyperparameters that should be predefined, and all signals share the same fixed thresholders.
Thus we transform soft thresholding operators into a trainable activation function so that the neural networks can learn a tailored thresholder matrix by back propagation with given tasks and datasets.
We initialize $\mathbf{\bar{S}}$ randomly to compute $\mathbf{S}$, and initialize $\bm{\theta}$ as a zero vector to generate $\mathbf{\Theta}$ in the beginning of training.
With these previous analyses, the $l$-th layer of the learnable GCN is formulated by
\begin{gather}\label{LGCN}
\begin{split}
\mathbf{H}^{(l)} = \sigma \left( \rho (\mathbf{A}_{s}) \mathbf{H}^{(l-1)} \mathbf{W}^{(l)}_{lgcn} \right),
\end{split}
\end{gather}
where $\mathbf{H}^{(0)} = \mathbf{H}$.
Namely, the trainable $\mathbf{H}$ obtained in the previous module becomes the unique common representation of multiple views and is regarded as the input features of nodes in GCN.
We use a widely employed 2-layer learnable GCN as an example, which computes the node embedding $\mathbf{Z}$ with
\begin{gather}\label{LGCN2Layer}
\begin{split}
\mathbf{Z} = \mathrm{softmax} \left(  \rho (\mathbf{A}_{s})  \sigma \left( \rho (\mathbf{A}_{s}) \mathbf{H} \mathbf{W}^{(1)}_{lgcn} \right) \mathbf{W}^{(2)}_{lgcn} \right).
\end{split}
\end{gather}
For a semi-supervised classification task, the loss function of learnable GCN is defined by the cross-entropy error over semi-supervised information generated from the labeled sample set $\Omega$, as shown below:
\begin{gather}\label{GCNLoss}
\begin{split}
\mathcal{L}_{lgcn} = - \sum_{i \in \Omega} \sum_{j=1}^{c} \mathbf{Y}_{ij} \mathrm{ln} \mathbf{Z}_{ij},
\end{split}
\end{gather}
where $\mathbf{Y} \in \mathbb{R}^{| \Omega | \times c}$ is the incomplete label matrix generated from $\Omega$ satisfying $| \Omega | \ll m$.

\begin{algorithm}[!tbp]
    \caption{Training Framework of LGCN-FF}
    \label{Algorithm}
    \textbf{Input}: Multi-view data $\mathcal{X} = \{\mathbf{X}^{(1)}, \cdots, \mathbf{X}^{(V)} \}$ and semi-supervised information $\mathbf{Y} \in \mathbb{R}^{| \Omega | \times c}$.\\
    \textbf{Output}: Node embedding $\mathbf{Z}$.
    
    \begin{algorithmic}[1]
    \STATE {Initialize weights and biases of sparse autoencoders;}
    \STATE {Initialize weights, biases and learnable input $\mathbf{H}$ of fully-connected networks;}
    \STATE {Initialize learnable weights $\{ \pi^{(v)} = \frac{1}{V}\}_{v=1}^{V}$, $\mathbf{S} \in \mathbb{R}^{m \times m}$ and $\mathbf{\Theta} \in \mathbb{R}^{m \times m}$ of learnable GCN;}
    \STATE {Initialize adjacency matrices $\mathbf{A}^{(1)}, \cdots, \mathbf{A}^{(V)}$ via KNN;}
    \WHILE {not convergent}
        \FOR {$v = 1 \rightarrow V$}
            \STATE {Compute $\mathbf{O}^{(\frac{L}{2}, v)}$ and $\mathbf{O}^{(L, v)}$ of the $v$-th sparse autoencoder with Equation \eqref{SAE};}
            \STATE {Update $\{ \mathbf{W}_{sa}^{(l, v)}, \mathbf{b}_{sa}^{(l, v)}\}_{l = 1}^{L}$ with back propagation;}
       \ENDFOR
    
       \FOR {$v = 1 \rightarrow V$}
            \STATE {Compute $\mathbf{G}^{(L)}$ of the fully-connected network with Equation \eqref{FCNet};}
       \ENDFOR
       \STATE {Update $\{ \mathbf{W}_{fc}^{(l)},  \mathbf{b}_{fc}^{(l)}\}_{l = 1}^{L}$ with back propagation;}
    
       \FOR {$v = 1 \rightarrow V$}
            \STATE {Compute $\mathbf{G}^{(L)}$ of the fully-connected network with Equation \eqref{FCNet};}
       \ENDFOR
       \STATE {Update $\mathbf{H}$ with back propagation;}
    
       \STATE {Compute $\mathbf{Z} = \mathbf{H}^{(L)}$ of learnable GCN with Equation \eqref{LGCN};}
       \STATE {Update $\{ \mathbf{W}_{lgcn}^{(l)} \}_{l=1}^{L}$, $\{\pi^{(v)}\}_{v=1}^{V}$, $\mathbf{S}$ and $\mathbf{\Theta}$ with back propagation;}
    \ENDWHILE
    \RETURN {Node embedding $\mathbf{Z}$.}
    \end{algorithmic}
\end{algorithm}

\subsection{Training Strategy}\label{TrainingStrategy}
The proposed LGCN-FF is an end-to-end neural network framework with a multi-step optimization method,
as described in Algorithm \ref{Algorithm}.
Because a single optimization may be not jointly convex for all variables,
we follow the ADM strategy \cite{LinLS11} and divide the optimization into the following four steps:
optimizing trainable weights and biases of sparse autoencoders, optimizing trainable weights and biases of the fully-connected network, optimizing the trainable input $\mathbf{H}$,
and optimizing trainable parameters in learnable GCN.
In an independent training iteration, each step performs one-step forward propagation,
and then conducts back propagation with fixed uncorrelated variables.
It is noticed that each step optimizes step-specific variables via its own loss function,
i.e., all sparse autoencoders in the first step employ Equation \eqref{SAELoss}, the second and the third steps share the same loss function defined in Equation \eqref{FCLoss},
and the final step applies Equation \eqref{GCNLoss}.
Although the formulated problem is optimized separately in the same iteration with the output of the former optimization becoming the input of the latter one,
the whole framework is organized by ADM strategy so that each convex subproblem can be solved effectively.
At each iteration, given multi-view data $\{\mathbf{X}^{(v)} \in \mathbb{R}^{m \times n} \}^{V}_{v=1}$ with $V$ views, the computational complexity for sparse autoencoders and the feature fusion network is $\mathcal{O}(2Vmnd + md^{2})$ if all embeddings are projected onto a $d$-dimension vector with $d \ll n$.
The forward propagation of learnable GCN costs $\mathcal{O}(mn + md^{2})$.

\section{Experimental Analyses}\label{Experiment}

\subsection{Experimental Settings}

\subsubsection{Datasets Description}
The proposed LGCN-FF framework is utilized to perform semi-supervised classification tasks on several real-world multi-view datasets.
Seven publicly available multi-view datasets are selected for performance evaluation, as listed below:
\begin{itemize}
  \item \textbf{ALOI\footnote{http://aloi.science.uva.nl}:} This is an image dataset which contains objects that are taken under varied light conditions or rotation angles. Multi-view features including 64-D RGB color histograms, 64-D HSV color histograms, 77-D color similarities and 13-D Haralick features are involved.
  \item \textbf{BBCnews\footnote{http://mlg.ucd.ie/datasets/segment.html}:} It is a collection of news reports which covers politics, entertainment, business, sport and technology fields.  There are totally 4 different textual features extracted from various segments to describe the news.
  \item \textbf{BBCsports\footnote{http://mlg.ucd.ie/datasets/bbc.html}:} Different from BBCnews, it is a dataset consisting of 5 different areas from BBC sport websites, including football, athletics, cricket, rugby and tennis news, illustrated from 2 distinct views.
  \item \textbf{MNIST\footnote{http://yann.lecun.com/exdb/mnist/}:} It is a well-known dataset of handwritten digits, where three types of features are extracted: 30-dimension IsoProjection, 9-dimension Linear Discriminant Analysis (LDA)  and 9-dimension Neighborhood Preserving Embedding (NPE) features.
  \item \textbf{Wikipedia\footnote{http://www.svcl.ucsd.edu/projects/crossmodal/}:} It is an article dataset that consists of 693 documents with 10 categories, which was crawled from Wikipedia website. Each entry is represented as two textual feature representations.
  \item \textbf{MSRC-v1\footnote{http://riemenschneider.hayko.at/vision/dataset/task.php?did=35}:} It is a well-known image dataset with totally 8 classes. Following previous work, a subset of this dataset with 7 classes is applied. There are five visual features extracted from each image in sum: 24-D color moment, 576-D  Histogram of Oriented Gradients (HOG), 512-D GIST, 256-D local binary pattern and 256-D CENTRIST features.
  \item \textbf{Reuters}\footnote{http://archive.ics.uci.edu/ml/machine-learning-databases/00259/}: This is a subset of Reuters dataset containing feature representations of documents that were written in five different languages: English, French, German, Spanish and Italian. Each language corresponds to an independent view.
\end{itemize}

A statistical summary of these datasets is presented in Table \ref{DataDescription}, including the numbers of views, features and classes.
\begin{table*}[!tbp]
    \centering
    \caption{A brief description of all test multi-view datasets.}
    \resizebox{\textwidth}{!}{
    \begin{tabular}{cccccc}
    \toprule
     Datasets&      \# Samples  & \# Views   & \# Features                  & \# Classes & Data Types\\
     \midrule
     ALOI           & 1,079     & 4          & 64 / 64 / 77 / 13                  & 10         & Object images\\
     BBCnews        & 685       & 4          & 4,659 / 4,633 / 4,665 / 4,684      & 5          & Textual documents\\
     BBCsports      & 544       & 2          & 3,183 / 3,203                  & 5          & Textual documents\\
     MNIST          & 10,000    & 3          & 30 / 9 / 9                       & 10         & Digit images\\
     Wikipedia      & 693       & 2          & 128 / 10                       & 10         & Textual documents\\
     MSRC-v1        & 210       & 5          & 24 / 576 / 512 / 256 / 254           & 7          & Object images\\
     Reuters        & 18,758    & 5          & 21,531 / 24,892 / 34,251 / 15,506 / 11,547           & 6          & Textual documents\\
    \bottomrule
    \end{tabular}}
    \label{DataDescription}
\end{table*}

\subsubsection{Compared Methods}
We compare the performance of the proposed LGCN-FF with following state-of-the-art methods:
\begin{itemize}
  \item \textbf{KNN:} $K$-Nearest Neighbor method is a classical non-parametric classification method looking for $k$ nearest training samples to conduct classification tasks.
  \item \textbf{AMGL:} Auto-weighted Multiple Graph Learning approach \cite{NieLiLi16Parameter} is a framework that adaptively learns weights for each view and explores a label indicator matrix by aggregating loss functions of numerous views. There is no extra hyperparameter in this method.
  \item \textbf{MVAR:} Multi-View semi-supervised classification algorithm via Adaptive Regression method \cite{TaoHNZY17} utilizes $\ell_{2,1}$-norm to calculate the regression loss value of each independent view, which constructs the objective function with the weighted sum of all regression losses.
  \item \textbf{MLAN:} Multi-view Learning with Adaptive Neighbors method \cite{NieCaiLi17Multiview} leverages a local structure embedding which is a unified framework conducting both unsupervised clustering and semi-supervised classification tasks.
  \item \textbf{AWDR:} Adaptive-Weighting Discriminative Regression approach \cite{YangDN19} is a multi-view classification algorithm where features from multifarious views are automatically assigned with the learned optimal weights.
  \item  \textbf{HLR-$\mathbf{M}^2$VS:} Hyper-Laplacian Regularized Multi-linear Multi-view Self-representations \cite{XieZQDT20} establishes a unified tensor space to jointly explore multi-view relationships via local geometrical structures, where a low-rank tensor regularization is adopted to guarantee that all views can come to an agreement.
  \item \textbf{ERL-MVSC:} Embedding Regularizer Learning for Multi-View
  Semi-supervised Classification \cite{HuangWZZL21} builds a framework integrating diversity, sparsity and consensus to flexibly address multi-view data, which projects a linear regression model to deduce view-specific embedding regularizers and automatically train weights of various perspectives. 
  \item \textbf{GCN fusion:} It is the well-known Graph Convolutional Network \cite{KipfW17} that deals with semi-supervised node classification tasks. Because the original model is not able to directly process multi-view data, we compute the average adjacency matrix during graph convolutions, which is designated as GCN fusion in this paper.
  \item \textbf{SSGCN fusion:} SSGCN \cite{ZhuK21} is a variant of GCN derived via a modified Markov diffusion kernel, which explores the global and local contexts of nodes. We also adopt an average weighted graph like GCN fusion.
  \item  \textbf{Co-GCN:} This is a GCN-based method \cite{LiLW20a} which adaptively employs the graph information from heterogeneous views with adaptive combined graph Laplacian matrices, which is optimized with a co-training strategy.
\end{itemize}

Most of these compared methods are graph-oriented algorithms, among which GCN fusion, SSGCN fusion and Co-GCN are based on GCN.
It is noted that Co-GCN is a state-of-the-art GCN-based framework for multi-view learning.
Actually only three GCN-based methods are involved in experiments, attributed to the fact that limited work has focused on GCN conducting downstream classification tasks with multi-view data.
This also amplifies the contribution of this work.

\subsubsection{Parameter Settings}
For most parameter settings, we follow the original settings of compared methods if feasible.
Note that AMGL is a parameter-free framework thus we do not need to predefine extra hyperparameters.
In particular, some parameter settings for compared methods are empirically set for better performance, as follows:
\begin{itemize}
  \item \textbf{KNN}: the number of neighbors is selected from $\{ 1, 3, 5, 7, 9\}$;
  \item \textbf{MVAR}: the trade-off weight for each view is tuned as $\lambda$ = 1000, and the redistribution parameter over views is set as $r$ = 2;
  \item \textbf{MLAN}: the number of adaptive neighbors is tuned in $[1,10]$;
  \item \textbf{AWDR}: the trade-off parameter is fixed as $\lambda=1.0$;
  \item \textbf{HLR-$\mathbf{M}^2$VS}: weighted factors are set as $\lambda_1=0.2$ and $\lambda_2=0.4$;
  \item \textbf{ERL-MVSC}: hyperparameters are set as $\alpha = 2$ and $\beta=\gamma=1$.
  \item \textbf{GCN fusion} and \textbf{SSGCN fusion}: a 2-layer GCN is employed and the learning rate is set as $0.001$;
  \item \textbf{Co-GCN}: the settings of the convolutional layers and learning rate are the same as those in GCN fusion.
\end{itemize}

As to LGCN-FF, we empirically adopt sparse autoencoders with the dimensions of latent representations selected from $\{256, 512, 1024, 2048\}$.
Adam optimizer is employed to update all learnable parameters with learning rate $lr=0.01$ for the feature fusion network and learnable GCN.
For all sparse autoencoders the learning rate is set to $lr = 0.001$.
We utilize $\ell_2$-norm as regularization for all learnable parameters and set weight decay as $0.01$.
Activation functions of learnable GCN and fully-connected network are set as $\mathrm{ReLU(\cdot)}$.
Sigmoid function is adopted as the activation function for sparse autoencoders.
Initial adjacency matrices are constructed by KNN.
Dropout rate of learnable GCN is set as $0.3$.
The default setting for hyperparameter controlling the sparsity penalty degree is $\beta = 1$.
Maximum number of iterations is set as 500.
In this paper, the proposed LGCN-FF framework is implemented by PyTorch platform and run on the machine with R9-5900X CPU, Nvidia RTX 3060 GPU and 32G RAM.
\begin{table*}[!tbp]
    \centering
    \caption{Classification accuracy (mean\% and standard deviation\%) of all compared semi-supervised classification methods with $10 \%$ labeled samples as supervision, where the best performance is highlighted in bold and the second best result is underlined. Limited by the computational complexity of algorithm and machine resources, some models encounter out-of-time or out-of-memory error on MNIST and Reuters datasets, marked with ``-".}
    \resizebox{\textwidth}{!}{
    \begin{tabular}{cccccccc}
      \toprule
          Datasets $\backslash$ Methods & ALOI & BBCnews & BBCsports & MNIST & Wikipedia & MSRC-v1 & Reuters \\ \midrule
          KNN & 45.7 (3.1) & 38.3 (9.6) & 38.3 (9.5) & 88.8 (0.4) & 58.7 (3.0) & 52.9 (8.8) & 34.4 (0.5) \\ 
          AMGL \cite{NieLiLi16Parameter} & 82.4 (3.3) & 52.3 (5.5) & 55.6 (1.4) & 88.5 (0.2) & 10.1 (0.8) & \underline{85.9 (1.8)} & - \\ 
          MVAR \cite{TaoHNZY17} & 72.9 (5.5) & 75.3 (5.5) & 83.7 (3.8) & 85.3 (0.8) & 61.2 (3.4) & 54.8 (7.5) & \underline{64.6 (0.3)} \\ 
          MLAN \cite{NieCaiLi17Multiview} & 87.6 (1.6) & 74.1 (0.9) & 62.6 (2.2) & 88.6 (0.3) & 10.2 (0.8) & 82.2 (5.4) & - \\ 
          AWDR \cite{YangDN19} & 93.6 (1.9) & 85.7 (1.4) & 81.3 (3.3) & 78.1 (0.3) & \underline{62.5 (5.8)} & 57.7 (7.0) & 61.3 (0.6) \\ 
          HLR-$\mathrm{M}^2$VS \cite{XieZQDT20} & 87.7 (1.7) & 78.1 (2.8) & 84.6 (0.4) & - & 36.5 (3.4) & 79.6 (8.4) & - \\ 
          ERL-MVSC \cite{HuangWZZL21} & 90.5 (2.7)  & 85.9 (2.2)  & 90.3 (1.9)  &  89.5 (0.3)  & 51.6 (2.3) & 73.3 (3.7)  &  - \\ \midrule
          GCN fusion \cite{KipfW17} & 92.6 (1.2) & 89.6 (1.8) & 87.0 (2.0) & 89.2 (0.5) & 60.1 (0.6) & 68.6 (7.2) & 55.4 (0.3) \\ 
          SSGCN fusion \cite{ZhuK21} & 93.4 (1.0) & \underline{89.7 (5.5)} & \underline{94.1 (1.2)} & 89.3 (0.1) & 62.1 (0.4) & 70.3 (4.9) & 56.8 (0.2) \\ 
          Co-GCN \cite{LiLW20a} & \underline{96.5 (0.4)} & 81.9 (1.5) & 84.8 (1.4) & \underline{89.9 (0.4)} & 57.9 (0.7) & 62.9 (4.4) & 60.2 (0.6) \\ \midrule
          LGCN-FF & \textbf{97.1 (0.6)} & \textbf{91.2 (0.9)} & \textbf{98.2 (0.6)} & \textbf{90.2 (0.3)} & \textbf{67.4 (1.5)} & \textbf{90.4 (2.0)} & \textbf{67.3 (0.5)} \\ \bottomrule
      \end{tabular}}
    \label{ACCcomparsionClassification}
\end{table*}

\begin{figure}[!tbp]
    \centering
    \includegraphics[width=\textwidth]{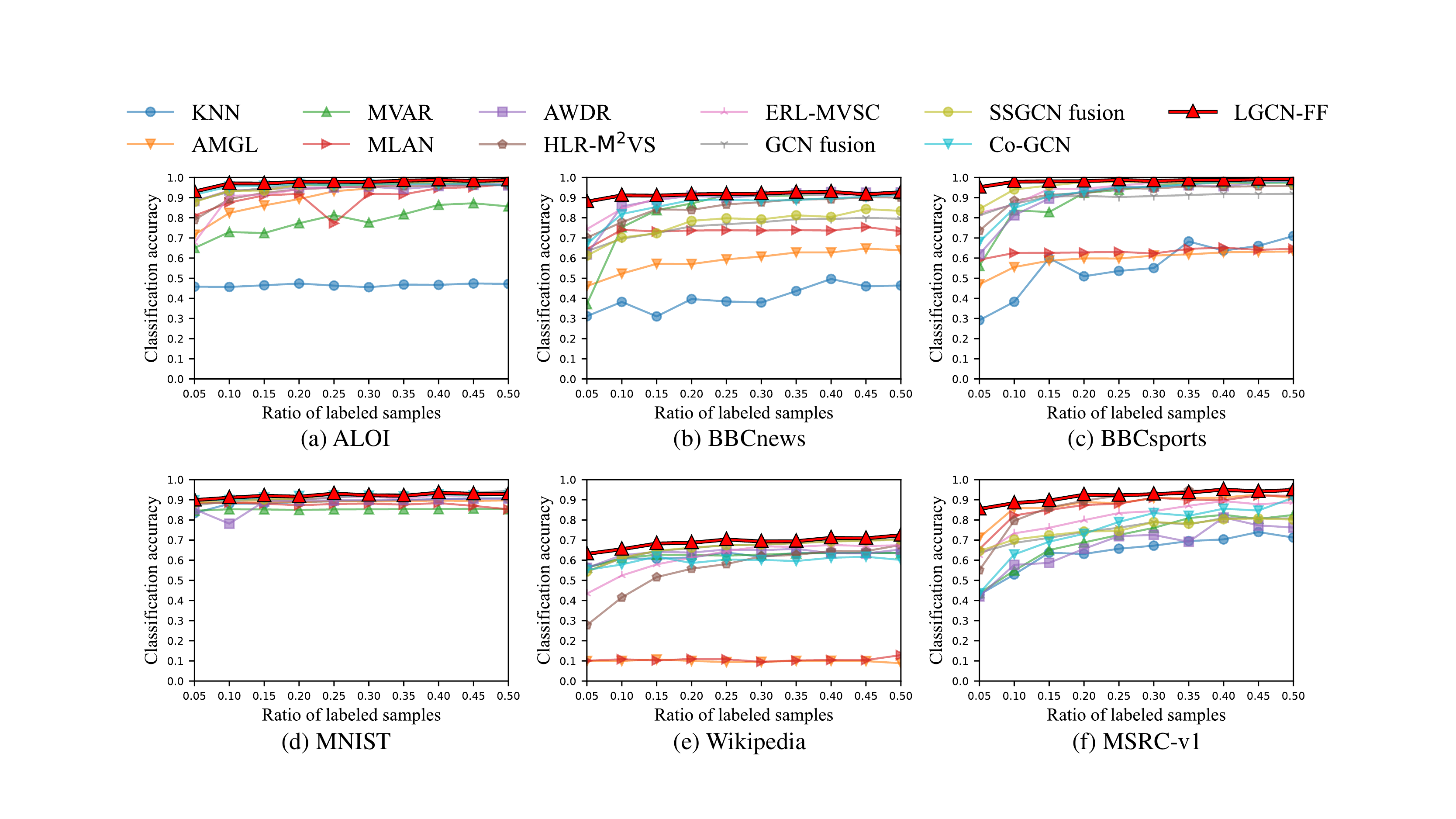}
    \caption{The varied performance of all compared methods as the ratio of labeled data ranges in $\{ 0.05, 0.10, \cdots, 0.50\}$ on ALOI, BBCnews, BBCsports, MNIST, Wikipedia and MSRC-v1 datasets.}
    \label{SemiRatio}
\end{figure}

\subsection{Semi-Supervised Classification}

\textbf{Classification Results:}
The performance of all compared methods with 10\% randomly labeled data is presented in Table \ref{ACCcomparsionClassification}, where the classification accuracy is used as an evaluation metric.
All methods are run 5 times and we record their average results and standard deviations.
We only compute cross-entropy errors $\mathcal{L}_{lgcn}$ of the learnable GCN under the supervision of 10\% labeled samples and evaluate the prediction performance with the rest $90\%$ unlabeled data.
The experimental results reveal that LGCN-FF reaches remarkable performance on all test datasets.
Compared with GCN-based methods, the performance improvement is more considerable on BBCnews, BBCsports, MSRC-v1 and Reuters datasets.
This observation suggests that the proposed LGCN-FF has stronger capacity of propagating node attributes among samples and extracting feature representations on relatively small datasets.
Besides, Figure \ref{SemiRatio} demonstrates the performance of all compared methods with various ratios of labeled samples.
The experimental results show that LGCN-FF performs satisfactorily with relatively small supervised ratios (e.g., 5\% or 10\% labeled samples) on all datasets,
and other algorithms generally require more supervised information to achieve comparable accuracy.
The performance improvement is more significant on BBCnews, BBCsports and MSRC-v1 datasets.
LGCN-FF also gains competitive accuracy with $5\%$ labeled samples on MNIST dataset, and outperforms other methods with more labels.
This indicates that LGCN-FF is more in line with the intention of semi-supervised classification.
In a nutshell, the proposed framework gains superior performance compared with these state-of-the-art approaches.
  
\begin{figure}[!tbp]
    \centering
    \includegraphics[width=0.67\textwidth]{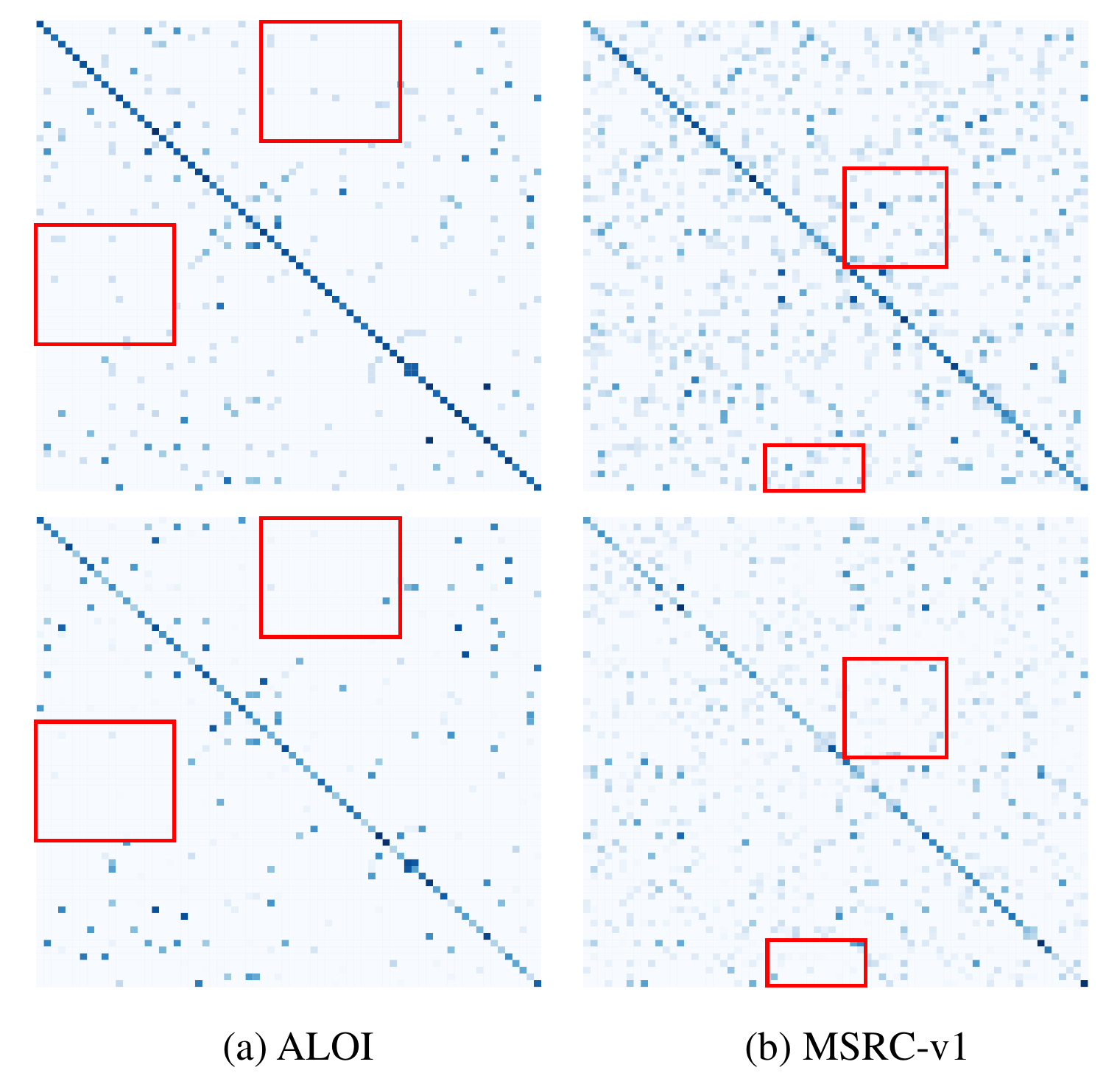}
    \caption{Visualization of weighted adjacency matrices (the first row) and the adjacency matrices learned by LGCN-FF (the second row) with selected datasets, where darker colors indicate higher element values of matrices.
    This figure only exhibits partial adjacency matrices for a better presentation, where red boxes highlight node connections that diminish or disappear.
    }
    \label{LearnedAdj}
\end{figure}

\textbf{Refined Adjacency Matrices:}
Figure \ref{LearnedAdj} presents the visualization of partial average weighted adjacency matrices and adjacency matrices learned by LGCN-FF.
Compared with a direct weighted sum strategy, the adjacency matrices refined by the DSA function are relatively pure.
It can be seen that some entries in learned adjacency matrices diminish or disappear, thereby resulting in sparser and more robust graphs.
The learned adjacency matrix makes critical node relationships more pronounced, which is beneficial for node embedding learning.
The pleasurable performance of LGCN-FF also favors the superiority of the proposed framework.

\textbf{Ablation Studies:}
In order to verify the effectiveness of the learnable GCN component, we also test the classification accuracy of the Weighted GCN-FF (WGCN-FF) that simply employs an average weighted adjacency matrix across all views.
Besides, the performance of Adaptive WGCN-FF (AWGCN-FF) is also recorded, where it learns weights of different adjacency matrices automatically and then directly utilizes graph convolution operations without further refining.
Actually, LGCN-FF is constructed based on AWGCN-FF, and adds a learnable DSA function $\rho(\cdot)$.
Results of the ablation study are presented in Table \ref{AblationStudy}.
It is worth mentioning that the performance of GCN fusion in Table \ref{ACCcomparsionClassification} can be regarded as the baseline accuracy.
It can be observed that LGCN-FF succeeds in promoting the performance of framework, which suggests the feasibility of the learnable GCN.
This may account for the reason that the learned adjacency matrix explores more discriminative relationships of nodes,
and reduces the impact of noises generated by different views.
Performance comparison also verifies that $\rho(\cdot)$ in LGCN-FF further promotes the accuracy of learning tasks.
\begin{table}[!htbp]
  \centering
  \caption{Ablation study of the proposed LGCN-FF on all test datasets, where average accuracy (\%) and standard deviation (\%) are recorded.}
  \begin{tabular}{cccc}
  \toprule
  Datasets $\backslash$ Methods &       WGCN-FF   &  AWGCN-FF       &  LGCN-FF          \\
   \midrule
   ALOI           & 91.8 (1.5)     & 94.1 (2.3)         & \textbf{97.1 (0.6)}                \\
   BBCnews        & 87.8 (1.1)     & 89.2 (1.3)         & \textbf{91.2 (0.9)}    \\
   BBCsports      & 96.5 (0.3)     & 96.9 (0.3)         & \textbf{98.2 (0.6)}               \\
   MNIST          & 88.0 (0.6)     & 88.8 (0.6)         & \textbf{90.1 (0.7)}                     \\
   Wikipedia      & 64.1 (1.5)     & 65.0 (1.6)         & \textbf{67.4 (1.5)}                    \\
   MSRC-v1        & 83.1 (2.4)     & 86.1 (2.3)         & \textbf{90.4 (2.0)}         \\
   Reuters        & 61.4 (0.6)     & 65.1 (0.4)         & \textbf{67.3 (0.5)}         \\
  \bottomrule
  \end{tabular}
  \label{AblationStudy}
\end{table}

\begin{figure}[!htbp]
  \centering
  \includegraphics[width=\textwidth]{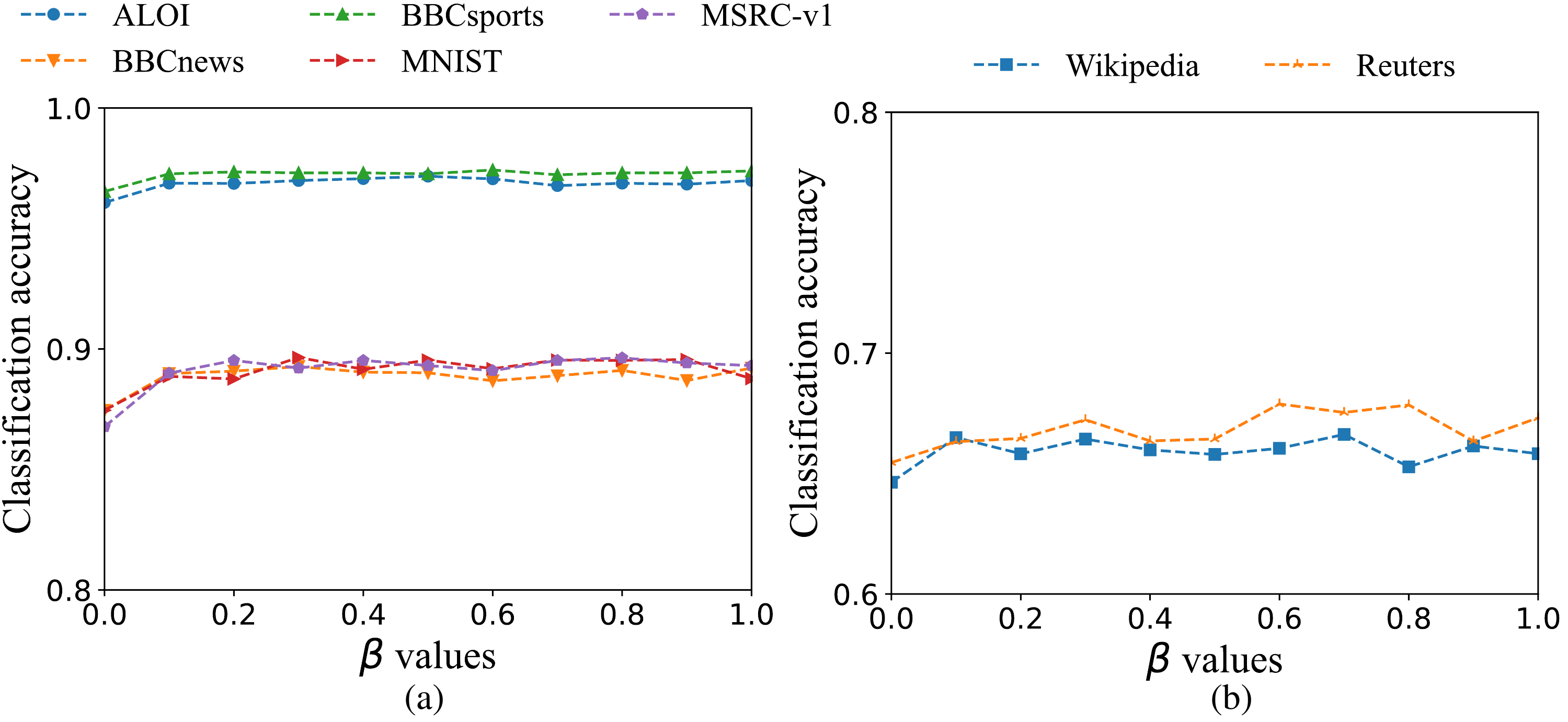}
  \caption{Classification accuracy of LGCN-FF with varying $\beta$ values on (a) ALOI, BBCsports, BBCnews, MSRC-v1 and MNIST, (b) Wikipedia and Reuters datasets.
  }
  \label{beta}
\end{figure}

\textbf{Impact of $\beta$:} Figure \ref{beta} analyzes the impact of $\beta$ with varying values, which is a coefficient controlling the sparsity penalty degree of sparse autoencoders in Equation \eqref{SAELoss}.
Experimental results reveal that the accuracy of LGCN-FF fluctuates slightly as $\beta$ changes on all datasets.
Nonetheless, it is observed that the accuracy declines marginally when $\beta$ decreases to 0.  
Namely, a vanilla autoencoder leads to undesired performance.
This indicates that suitable sparseness is beneficial for exploring hidden representations via sparse autoencoders.

\textbf{Convergence Analyses:}
\begin{figure*}[!tbp]
    \centering
    \includegraphics[width=\textwidth]{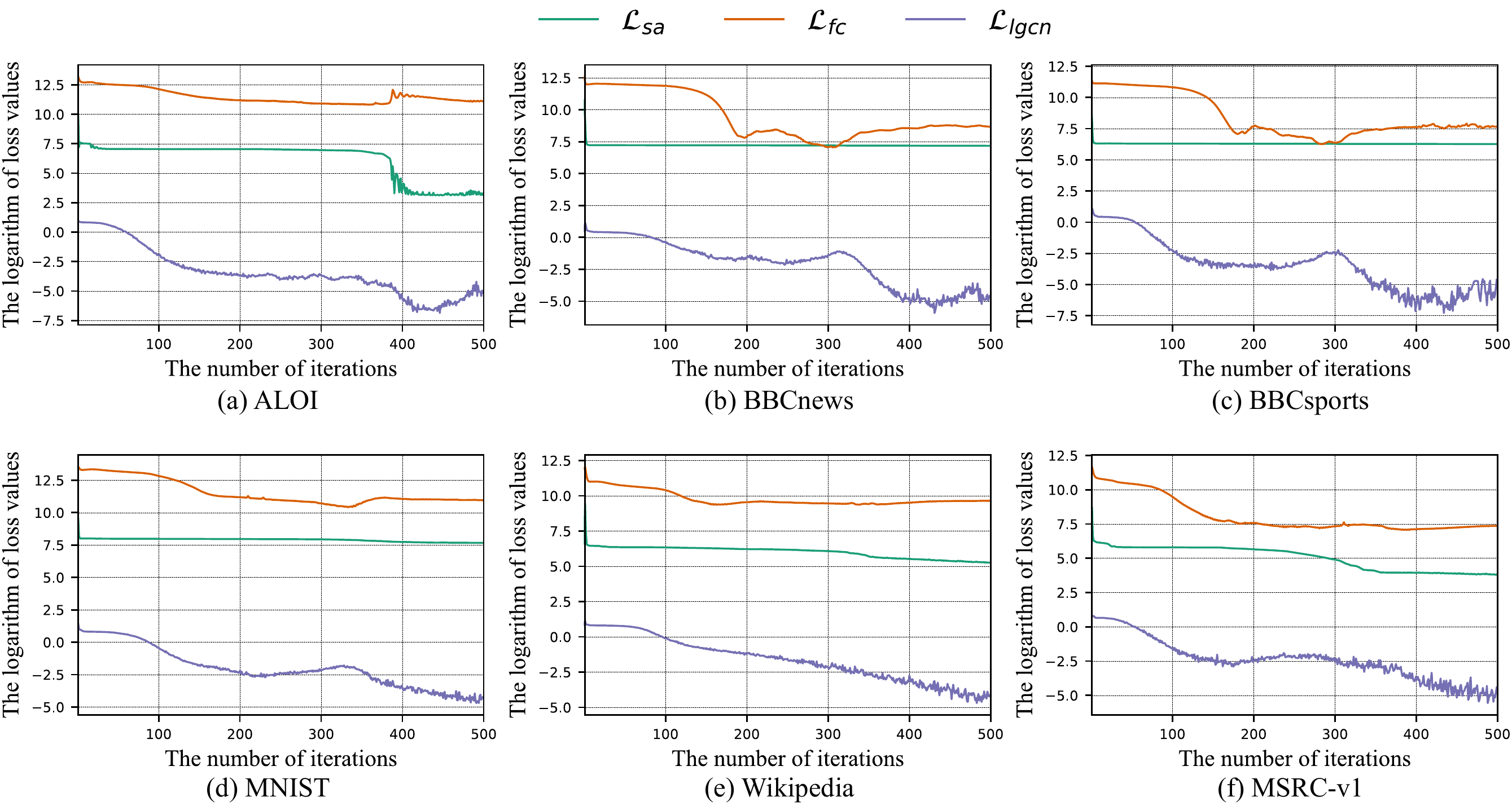}
    \caption{Convergence curves of logarithm loss values measured by $\mathcal{L}_{sa}$, $\mathcal{L}_{fc}$ and $\mathcal{L}_{lgcn}$.
    }
    \label{convergence}
\end{figure*}
Figure \ref{convergence} shows the convergence of the proposed LGCN-FF.
Because there are three loss functions in the framework, we plot logarithm loss values in a single figure for a better presentation.
For simplicity, we draw curves of $\mathcal{L}_{sa} = \sum_{v=1}^{V} \mathcal{L}_{sa}^{(v)}$.
From all subfigures in Figure \ref{convergence}, we have the following beneficial observations.
First of all, the loss of $\mathcal{L}_{sa}$ generally declines dramatically within 10 iterations, because of which the curve of $\mathcal{L}_{sa}$ tends to be a vertical line in the beginning of training.
Then it converges slightly on most datasets except ALOI, where the loss value drops considerably around 380 iterations.
Second, the loss value of $\mathcal{L}_{fc}$ starts to decrease significantly after a period of training (after 100 iterations in most cases),
attributed to the fact that the feature fusion network learns discriminative fusion features when the latent embeddings learned by sparse autoencoders are relatively fixed.
It is notable that the feature fusion network aims to seek a trade-off among multiple views, because of which the loss value may be relatively higher.
Third, there is an interesting phenomenon that the value of  $\mathcal{L}_{fc}$ may bounce marginally.
On the contrary, the loss of $\mathcal{L}_{lgcn}$ may plunge almost in the same time on some datasets.
This observation is more significant on ALOI (around 380 iterations), BBCnews (around 320 iterations), BBCsports (around 300 iterations) and MNIST (around 330 iterations).
One reasonable explanation is that the proposed collaborative training procedure allows the feature fusion network to refine shared embeddings with promising generalization capacity via flexible optimization, thereby leading to further improvement on the accuracy of downstream tasks.
Although the value of $\mathcal{L}_{fc}$ may not reach the lowest point, the learned feature fusion is a better trade-off among multifarious views.
Last but not the least, the value of $\mathcal{L}_{lgcn}$, which is directly related to the performance of downstream classification tasks, reaches the lowest point within 500 iterations on all datasets. 
The values of $\mathcal{L}_{lgcn}$ may fluctuate in the late period of training, owing to the varying features generated by the previous feature fusion network.
Because the classification accuracy of GCN is tightly related to the input features, the cross entropy loss is sensitive to the quality of feature fusion.
However, $\mathcal{L}_{lgcn}$ generally converges and it fluctuates in a small range.
It is noted that the fluctuation of $\mathcal{L}_{lgcn}$ may be amplified when it is within $[0, 1]$, because we adopt the logarithm of loss values for better presentation.
Actually, the fluctuation is marginal.
We can terminate the network training early when the value of loss $\mathcal{L}_{lgcn}$ does not continue to drop for several iterations.

\section{Conclusion}\label{Conclusion}
In this paper, we proposed an end-to-end neural network framework dubbed LGCN-FF which solved the multi-view learning problem with a learnable GCN and feature fusion network.
In feature fusion networks, multiple sparse autoencoders and a fully-connected network were utilized to fuse features from different views and study a unique underlying representation containing characteristics from all views.
The graph fusion procedure was conducted by the learnable GCN that adaptively integrated multiple topology graphs from multifarious views.
In addition, a learnable DSA function was proposed to learn a more robust shared adjacency matrix, which promoted the performance of LGCN-FF.
Finally, the proposed framework divided the optimization target into several subproblems and jointly learned feature and graph fusion representations with a multi-step optimization strategy.
Experimental results validated the superiority of the proposed framework in terms of multi-view semi-supervised classification tasks.

Here remain several interesting potential research directions to be further explored.
Existing GCN-based methods usually concentrate on undirected graphs, while node relationships in real-world applications are more likely to be directed graphs.
In most real-world applications, there is no natural topology network for most data and most graph information is established via the KNN algorithm.
It would be helpful if a new graph learning framework is developed.
In the future, we will devote more effort to feasible graph fusion learning with multi-view data.

\section*{Acknowledgments}
This work was partially supported by the National Natural Science Foundation of China (Nos. U21A20472 and 61672159).

\bibliographystyle{elsarticle-num}
\bibliography{LGCN}

\end{document}